%% file: ms.tex
\documentclass{article}
\usepackage[preprint, nonatbib]{neurips_2020}
\pdfoutput=1
\usepackage[utf8]{inputenc} 
\usepackage[T1]{fontenc}    
\usepackage{hyperref}       
\hypersetup{pdfauthor={Sharma, Singh, Devi, Murty},pdftitle={CATSETMAT}}
\usepackage{url}            
\usepackage{booktabs}       
\usepackage{amsfonts}       
\usepackage{nicefrac}       
\usepackage{microtype}      
\usepackage[numbers]{natbib}
\usepackage{hyperref}
\usepackage{subfig}
\usepackage{wrapfig}
\usepackage{graphicx}
\usepackage{booktabs}
\usepackage{xcolor}
\usepackage{amsmath}
\usepackage{amssymb}
\usepackage{amsthm}
\usepackage{wrapfig}
\usepackage{bm}

\newtheorem{definition}{Definition}

\newtheorem{lemma}{Lemma}
\newtheorem{observation}{Observation}

\newtheorem{note}{Note}

\newcommand{\dataicc}{\texttt{tmdb-cc}}
\newcommand{\dataick}{\texttt{tmdb-ck}}
\newcommand{\datamacm}{\texttt{mag-acm-ak}}

\newcommand{\algoncme}{\texttt{n2v-cross-mean}}
\newcommand{\algonfme}{\texttt{n2v-full-mean}}
\newcommand{\algoncmi}{\texttt{n2v-cross-min}}
\newcommand{\algonfmi}{\texttt{n2v-full-min}}

\newcommand{\algofsp}{\texttt{FSPool}}

\newcommand{\algohsnn}{\texttt{Hyper-SAGNN}}

\newcommand{\algocsm}{\texttt{CATSETMAT}}
\newcommand{\algocsmx}{\texttt{CATSETMAT-X}}
\newcommand{\algocsmsx}{\texttt{CATSETMAT-SX}}
\newcommand{\algocsmsxs}{\texttt{CATSETMAT-SXS}}

\newcommand{\algolpminaamean}{\texttt{bipartite-AA-min}}
\newcommand{\algolpmincnmean}{\texttt{bipartite-CN-min}}
\newcommand{\algolpmaxaamean}{\texttt{bipartite-AA-max}}
\newcommand{\algolpmaxcnmean}{\texttt{bipartite-CN-max}}
\newcommand{\algolpavgaamean}{\texttt{bipartite-AA-avg}}
\newcommand{\algolpavgcnmean}{\texttt{bipartite-CN-avg}}

\newcommand{\sat}{\textsl{\textsf{SAT}}}
\newcommand{\satp}{\textsl{\textsf{SAT}${}^\prime$}}
\newcommand{\cat}{\textsl{\textsf{CAT}}}
\newcommand{\catsetmat}{{\textit{CATSETMAT}}}

\newcommand{\V}{\mathcal{V}}
\newcommand{\F}{\mathcal{F}}
\newcommand{\B}{\mathcal{B}}
\renewcommand{\H}{\mathcal{H}}
\renewcommand{\P}{\mathcal{P}}
\newcommand{\N}{\mathcal{N}}
\renewcommand{\S}{\mathcal{S}}
\newcommand{\D}{\mathcal{D}}
\newcommand{\X}{\mathcal{X}}
\newcommand{\Y}{\mathcal{Y}}
\newcommand{\M}{\mathcal{M}}
\newcommand{\E}{\mathcal{E}}
\newcommand{\G}{\mathcal{G}}

\newcommand{\x}{\mathbf{x}}
\newcommand{\y}{\mathbf{y}}
\renewcommand{\d}{\mathbf{d}}
\newcommand{\s}{\mathbf{s}}
\newcommand{\deltabf}{\bm{\delta}}
\newcommand{\W}{\mathbf{W}}
\newcommand{\cross}{\times}

\newenvironment{myquote}[1]%
  {\list{}{\leftmargin=#1\rightmargin=#1}\item[]}%
  {\endlist}

\title{The CAT SET on the MAT: Cross Attention for Set Matching in Bipartite Hypergraphs}
\vspace{-2cm}
\author{%
  Govind Sharma \qquad Swyam Prakash Singh \qquad V. Susheela Devi \qquad M. Narasimha Murty\\~\\
  \{\texttt{\href{mailto:govinds@iisc.ac.in}{govinds}, \href{mailto:swyamsingh@iisc.ac.in}{swyamsingh}, \href{mailto:susheela@iisc.ac.in}{susheela}, \href{mailto:mnm@iisc.ac.in}{mnm}\}@iisc.ac.in}\\~\\
  Department of Computer Science and Automation\\
  Indian Institute of Science, Bangalore\\
  Karnataka 560012, India
}

\begin{document}
\maketitle
    \input{sections/abstract}

    \section{Introduction}
        \label{sec:intro}
        \input{sections/intro}
    
    
    \section{Bipartite Hyperedge Prediction}
        \label{sec:bipred}
        \input{sections/bipred}
    
    \section{Bipartite Hyperedge Set Matching Prediction (BHSMP)}
        \label{sec:setmat}
        \input{sections/setmat}

    \section{The Cross Attention (CAT) Framework}
        \label{sec:cat}
        \input{sections/cat}
    
    \section{The \catsetmat{} Architecture}
        \label{sec:catsetmat}
        \input{sections/catsetmat}

    \section{Related Work}
        \label{sec:relwork}
        \input{sections/relwork}

    \section{Experiments}
        \label{sec:exp}
        \input{sections/exp}
    
    \section{Results and Discussion}
        \label{sec:results}
        \input{sections/results}

    \section{Conclusion and Future Work}
        \label{sec:conc}
        \input{sections/conc}
    
    \section*{Broader Impact (As required by NeurIPS)}
    The focus of our research is a special higher-order structure that shows up in the real-world more often than expected.
    Co-morbidity, drug-abuse warning, complex chemical reactions, etc. all involve a higher-order bipartite relational structure.
    Not-to-mention, other areas such as entertainment, meta-research, fashion-technology, \textit{etc.} also exhibit bipartite higher-order relationships.
    Never in the past have these hypergraphs been analyzed with the lens of applicability, let alone from the perspective of machine learning.
    Our work is the first of the hopefully many forthcoming pieces of research that explore bipartite hypergraphs in such detail.
    Long have networks been modeled as standard pairwise-edged graphs; the interest in higher-order relations is increasing rapidly, which would call for such research areas to be explored for novel applications of the same.
    
\bibliography{ms}
\bibliographystyle{plain}

\appendix
\input{sections/supplementary_content}
\end{document}

%% file: sections/abstract.tex
\begin{abstract}

Usual relations between entities could be captured using graphs; but those of a higher-order -- more so between two different types of entities (which we term "left" and "right") -- calls for a "bipartite hypergraph". For example, given a left set of symptoms and right set of diseases, the relation between a set subset of symptoms (that a patient experiences at a given point of time) and a subset of diseases (that he/she might be diagnosed with) could be well-represented using a bipartite hyperedge. The state-of-the-art in embedding nodes of a hypergraph is based on learning the self-attention structure between node-pairs from a hyperedge. In the present work, given a bipartite hypergraph, we aim at capturing relations between node pairs from the cross-product between the left and right hyperedges, and term it a "cross-attention" (CAT) based model. More precisely, we pose "bipartite hyperedge link prediction" as a set-matching (SETMAT) problem and propose a novel neural network architecture called CATSETMAT for the same. We perform extensive experiments on multiple bipartite hypergraph datasets to show the superior performance of CATSETMAT, which we compare with multiple techniques from the state-of-the-art. Our results also elucidate information flow in self- and cross-attention scenarios.
\end{abstract}


%% file: sections/intro.tex
Relations between two entities are easily captured by a \textit{graph}~\cite{newman2006structure,newman2003structure}, wherein a collection of pairwise \textit{edges} (either directed or undirected) encapsulates the relational structure (\textit{e.g.}, friendship relations between two people on a social network~\cite{traud2012social,adamic2003friends}).
Moreover, \textit{heterogeneous graphs}~\cite{sun2012mining,wang2019heterogeneous} are used to capture relationship structures between entities of multiple ``types'' (\textit{e.g.}, a bibliographic network~\cite{deng2012modeling} between nodes of type author, paper, venue, \textit{etc.}).
However, when the number of types is restricted to two (say, ``left'' and ``right''), and relations exist only \textit{across} (and not \textit{among}) them, we resort to a \textit{bipartite graph}~\cite{kitsak2017latent} (\textit{e.g.}, an author-paper bibliography network).

Nevertheless, any such relation captured by a usual network --- be it homogeneous, heterogeneous, or bipartite --- is strictly restricted to a \textit{pair} of entities.
But relations in nature, more often than not, occur between more than two entities.
For example, a co-authorship network (wherein usually, a relation is said to exist between a pair of authors who have co-authored at least one article) is, in fact, a network where possibly more than two authors (all those who have co-authored at least one article) can be connected via a single higher-order relation.
A collection of such higher-order relations (\textit{hyperedges}) is called a \textit{hypergraph}~\cite{bretto2013hypergraph,berge1973graphs}, and is used since using a graph for the job proves to be lossy~\cite{zhou2007learning}.
Akin to graphs, hypergraphs too have their own heterogeneous versions (those that capture higher-order relations between nodes of different types), which have been used in the literature to capture relations of the type buyer-broker-seller~\cite{bonacich2004hyper}, user-location~\cite{yang2019revisiting}, \textit{etc}.
\begin{wrapfigure}{r}{0.42\textwidth}
    \centering
    \vspace{-0.25cm}
    \includegraphics[width=0.4\textwidth]{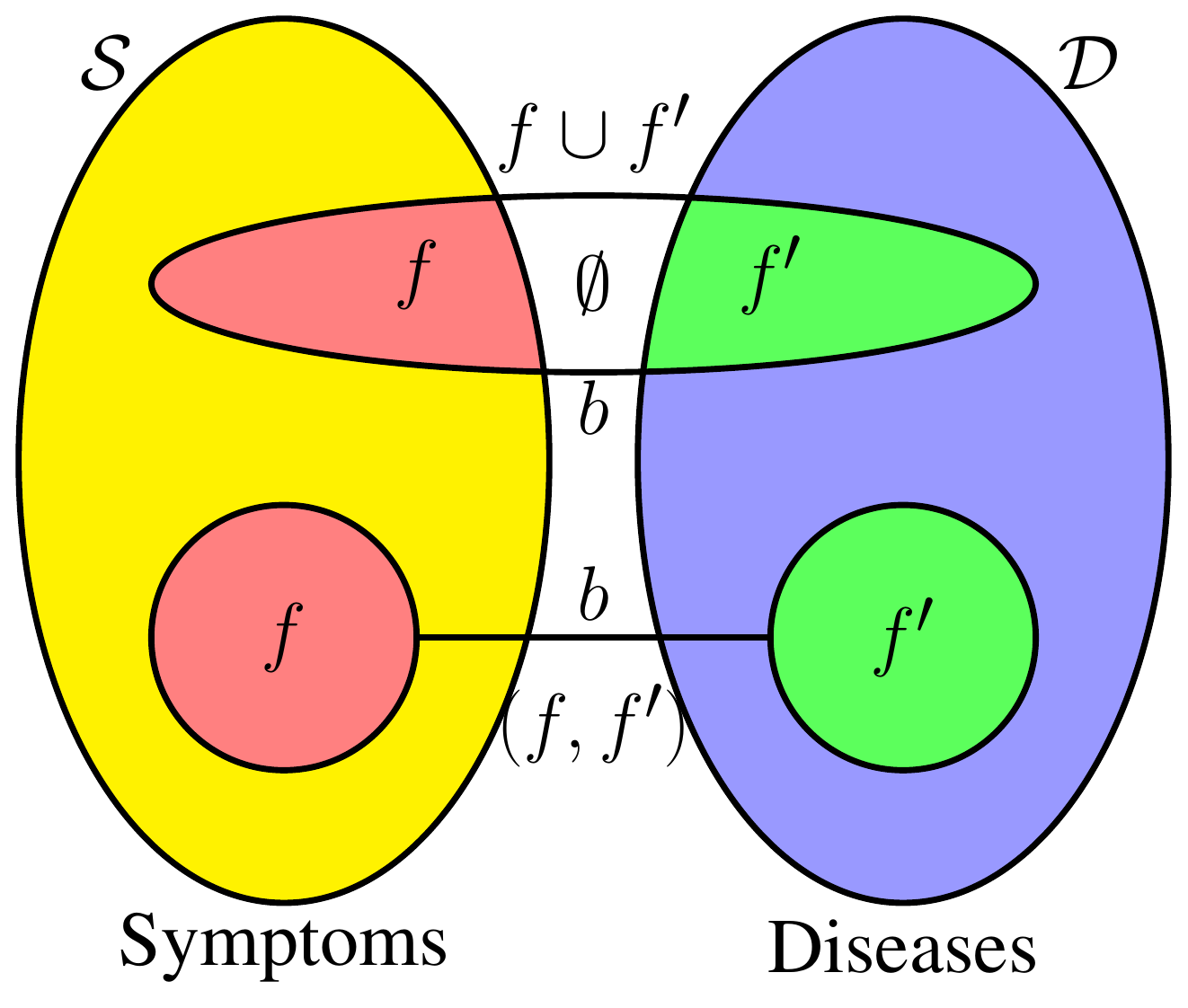}
    \caption{Example of a \textit{symptoms-diseases} bipartite hypergraph with left and right node sets being those of symptoms $\S$ and diseases $\D$.
    Higher-order bipartite relations could be interpreted as a single bipartite hyperedge (above) $b = f \cup f^\prime \in 2^{\S \cup \D}$ intersecting the node sets at $f$ and $f^\prime$.
    Such a relation can also be viewed (below) as a set match $(f, f^\prime) \in 2^\S \times 2^\D$ which comes as a result of a set matching between symptoms and diseases.}
    \vspace{-0.4cm}
    \label{fig:disease-symptom}
\end{wrapfigure}
While much has been done about such \textit{heterogeneous hypergraphs} in the literature~\cite{gui2016large,zhu2016heterogeneous}, \textit{bipartite hypergraphs}~\cite{zverovich2006bipartite} --- hypergraphs wherein each hyperedge is required to have at least one node from each one of two disjoint node-sets ``left'' and ``right'' --- have \textit{seldom} been talked about.


An example would be the \textit{symptoms-diseases} bipartite hypergraph shown in Fig.~\ref{fig:disease-symptom}, where given a (left) set $\S$ of symptoms and another (right) set $\D$ of diseases, every \textit{non-trivial diagnosis} --- one wherein the doctor identifies at least one symptom (\textit{i.e.}, $f \in 2^\S \setminus \emptyset$) in a patient and diagnoses him/her with at least one disease (\textit{i.e.}, $f^\prime \in 2^\D \setminus \emptyset$) --- forms a bipartite hyperedge $f \cup f^\prime$, and a collection $\B$ of such hyperedges forms a bipartite hypergraph $\H = (\S, \D, \B)$.
For the diagnosis $f$ denotes all the its symptoms and $f^\prime$ denotes all diseases that the patient is suffering from.

Existing models use a self-attention mechanism (Hyper-SAGNN) to predict a heterogeneous hyperedge, but miss the fact that the flow of information has to be across the right and left sets, and not among them individually.
In this work we too aim to learn a neural network model for predicting relations, only in a bipartite hypergraph.
Going by the symptoms and diseases example in Fig.~\ref{fig:disease-symptom}, we need not model the existence of a set of symptoms or a set of diseases, but the existence of a relation between a set of symptoms and a set of diseases.

We repose hyperedge prediction as a \textit{set-matching} (SETMAT) problem: \textit{given two set-of-sets, what pair of sets ``match'' with each other}?
In our case, the two sets would be the left and right hyperedges, and we call it a ``match'' if they are linked by a bipartite hyperedge.

The following is a list of all the contributions we make in this work:
\begin{enumerate}
    \item This is the very first work on bipartite hypergraphs in machine learning, along with introduction of some novel \textbf{datasets}.
    Moreover, we introduce the problem of \textbf{predicting higher-order bipartite relations} in networks for the first time.
    \item Elucidate the drawback of usual hyperedge embedding techniques for bipartite hyperedges via a \textbf{alternating positive/negative set pairs} based explanation.
    \item Pose the above problem as a \textbf{set-matching} prediction problem and show theoretical equivalence of the same.
    \item Formulate a \textbf{cross-attention framework} based neural network architecture to deal with the set-matching and hence the bipartite hyperedge prediction problem.
\end{enumerate}

We make our code\footnote{\url{https://github.com/govindjsk/catsetmat}} and all our datasets publicly available online; please refer to Section~\ref{app:exp:data_desc} for how to access the datasets.

%% file: sections/bipred.tex
\subsection{Bipartite Hypergraphs}
Given are two sets of disjoint nodes: \textbf{left nodes} $\V$ and \textbf{right nodes} $\V^\prime$; a \textbf{simple bipartite hypergraph} is defined as $\H_{sim} = (\V, \V^\prime, \B)$, where $\B \subseteq 2^{\V \cup \V^\prime}$ such that there is at least one node from each node set in the elements of $\B$.
However, we proceed further and define a different kind of bipartite hypergraph: a \textbf{per-fixed}\footnote{We call it ``per-fixed'' and not pre-fixed, since only left and right hyperedges are fixed.} one wherein basically we \textit{fix} the set of left and right hyperedges beforehand.
Over the node sets $\V$  and $\V^\prime$, the set $\mathbb{F} := 2^\V \setminus \emptyset$ of \textit{potential left hyperedges} and the set $\mathbb{F}^\prime := 2^{\V^\prime} \setminus \emptyset$ of \textit{potential right hyperedges} could be noted.
Once we \textit{fix} the set of \textit{actual} \textbf{left hyperedges} to be $\F \subseteq \mathbb{F}$ and \textit{actual} \textbf{right hyperedges} to be $\F^\prime \subseteq \mathbb{F}^\prime$, we could define the set of \textit{potential bipartite hyperedges} as:
\begin{equation}
    \mathbb{B}(\F, \F^\prime) := \{f\cup f^\prime \mid f \in \F, f^\prime \in \F^\prime\}.
\end{equation}
\begin{definition}[\textbf{\textsc{Per-fixed Bipartite Hypergraph}}]
\label{def:bipartite_hypergraph}
A per-fixed bipartite hypergraph is an ordered set $\H = (\V, \V^\prime, \F, \F^\prime, \B)$ of left vertices $\V$, right vertices $\V^\prime$, fixed left hyperedges $\F$, fixed right hyperedges $\F^\prime$, and \textbf{bipartite hyperedges} $\B \subseteq \mathbb{B}(\F, \F^\prime)$.
Furthermore, $\hat{\B} := \mathbb{B}(\F, \F^\prime) \setminus \B$ denotes the set of all \textbf{bipartite non-hyperedges}.
\end{definition}
A simple bipartite hypergraph is different from a usual (non-bipartite) hypergraph in that it has two disjoint sets of nodes ($\V, \V^\prime$) instead of one.
On the other hand, a per-fixed bipartite hypergraph is different from the two, since we fix the sets of left and right hyperedges ($\F, \F^\prime$) beforehand, and are thence worried only about connections across them (as defined by $\B$).
An important consequence of these facts is that \textbf{we need not model the existence of the left or right hyperedges individually, but focus on the cross bipartite relations instead}.

\begin{observation}
\label{obs:hyg_as_graph}
Given a per-fixed bipartite hypergraph $\H = (\V, \V^\prime, \F, \F^\prime, \B)$, the triplet $\G := (\F, \F^\prime, \B)$ forms a bipartite graph over node sets $\F$ and $\F^\prime$.
Also, the triplet $\H_{sim} := (\V, \V^\prime, \B)$ forms a simple bipartite hypergraph.
\end{observation}

\begin{note}
Henceforth, unless prefixed with the term ``simple'', \textbf{the phrase ``bipartite hypergraph'' would refer to a per-fixed bipartite hypergraph} as per Definition~\ref{def:bipartite_hypergraph}.
\end{note}

\subsection{The Bipartite Hyperedge Prediction Problem}
In the present work, we have set out to solve the problem of bipartite hyperedge prediction (BHP), which we define as follows:
\begin{definition}[\textbf{Bipartite Hyperedge Prediction (BHP)}]
\label{def:bhp}
Given $\H = (\V, \V^\prime, \F, \F^\prime, \B)$, learn a \textbf{BHP predictor} $\varphi: \mathbb{B}(\F, \F^\prime) \rightarrow \mathbb{R}$ such that for $f, \hat{f} \in \F$ and $f^\prime, \hat{f}^\prime \in \F^\prime$ and disjoint sets $\P \subseteq \B$ and $\N \subseteq \hat{\B}$ (see Def.~\ref{def:bipartite_hypergraph}) denoting the \textbf{positive} and the \textbf{negative class} respectively, we have:
\begin{equation}
\label{eq:bhcp}
    \max_{\varphi \in \mathbb{R}^{\mathbb{B}(\F, \F^\prime)}} Pr(\varphi(f \cup f^\prime) > \varphi(\hat{f} \cup \hat{f}^\prime) \mid f \cup f^\prime \in \P \text{ and } \hat{f} \cup \hat{f}^\prime \in \N)
\end{equation}
\end{definition}
It is to be noted that Definition~\ref{def:bhp} considers a per-fixed bipartite hypergraph and one could also define the BHP problem for a simple bipartite hypergraph (let's call it simple-BHP) as well.
The only difference between BHP and simple-BHP would be that while for the former, the set of possible left- and right-hyperedges is fixed even before the problem is defined, for the latter, any possible left- or right-hyperedge could be chosen as arbitrary subset choices from the respective node sets.
Nevertheless, the difference shows up only for bipartite non-hyperedges: for simple-BHP, we have $\B \subseteq \mathbb{B}(\mathbb{F}, \mathbb{F}^\prime)$ and $\hat\B := \mathbb{B}(\mathbb{F}, \mathbb{F}^\prime) \setminus \B$, but for per-fixed BHP (as per Definition~\ref{def:bhp}, the set of observed left hyperedges and right hyperedges are fixed to $\F \subseteq \mathbb{F}$ and $\F^\prime \subseteq \mathbb{F}^\prime$ respectively, and then hyperedges and non-hypergedges are defined as $\B \subseteq \mathbb{B}(\F, \F^\prime)$ and $\hat\B := \mathbb{B}(\F, \F^\prime) \setminus \B$.
\textbf{We will see how existing hyperege prediction algorithms apply only to simple-BHP and fail to cater the per-fixed BHP problem}.



%% file: sections/setmat.tex
\subsection{Set Matching (SETMAT)}
\begin{definition}[\textbf{Set Matching}]
\label{def:set_matching}
Given two sets-of-sets $\X$ and $\Y$, a \textbf{set matching} is defined as a relation $\M \subseteq \X \times \Y$ that matches a set element $X \in \X$ to another set element $Y \in \Y$.
Every $(X, Y) \in \M$ is called a \textbf{set match} (or \textbf{match} for short).
Naturally, the set $\hat{\M} := \X \times \Y \setminus \M$ refers to the corresponding \textbf{set anti-matching} and an element $(\hat{X}, \hat{Y}) \in \hat{\M}$ is called a \textbf{set anti-match} (or \textbf{anti-match} or \textit{non-match} for short).
\end{definition}

\begin{definition}[\textbf{Set Matching Predicton (SMP)}]
Given two sets-of-sets $\X$ and $\Y$, and a set matching $\M \subseteq \X \times \Y$, the set matching prediction problem learns an \textbf{SMP predictor} $\mu: \X \times \Y \rightarrow \mathbb{R}$ such that for $X, \hat{X} \in \X$ and $Y, \hat{Y} \in \Y$, we have:
\begin{equation}
\label{eq:smp}
    (X, Y) \in \M \text{ and } (\hat{X}, \hat{Y}) \in \hat{\M} 
    \implies \mu(X, Y) \geq \mu(\hat{X}, \hat{Y}).
\end{equation}
\end{definition}

\begin{lemma}
Every bipartite hypergraph is a set-matching over its left and right hyperedges.\footnote{Proof in Appendix~\ref{app:proofs:lemma1}.}
\end{lemma}

\subsection{BHP as SMP}
\begin{lemma}
The BHP problem for a bipartite hypergraph can be solved by an SMP predictor for its equivalent set-matching.\footnote{Proof in Appendix~\ref{app:proofs:lemma2}.}
\end{lemma}

The foregoing lemma states an important result: \textbf{that we could solve the BHP problem using SMP}.
Since we have an equivalent set matching for a given hypergraph, and that we could solve the BHCP problem using SMP, let us define a new problem called the \textbf{Bipartite Hyperedge Set Matching Prediction (BHSMP)} problem as follows:
\begin{definition}[\textbf{Bipartite Hyperedge Set Matching Prediction (BHSMP)}]
\label{def:smbhp}
Given a bipartite hypergraph $\H = (\V, \V^\prime, \F, \F^\prime, \B)$, a \textbf{BHSMP predictor} is simply defined as an SMP predictor for the equivalent set matching.
\end{definition}

%% file: sections/cat.tex
\subsection{The Problem with usual Self-Attention}
\begin{figure}[htb]
    \centering
    \includegraphics[width=0.6\textwidth]{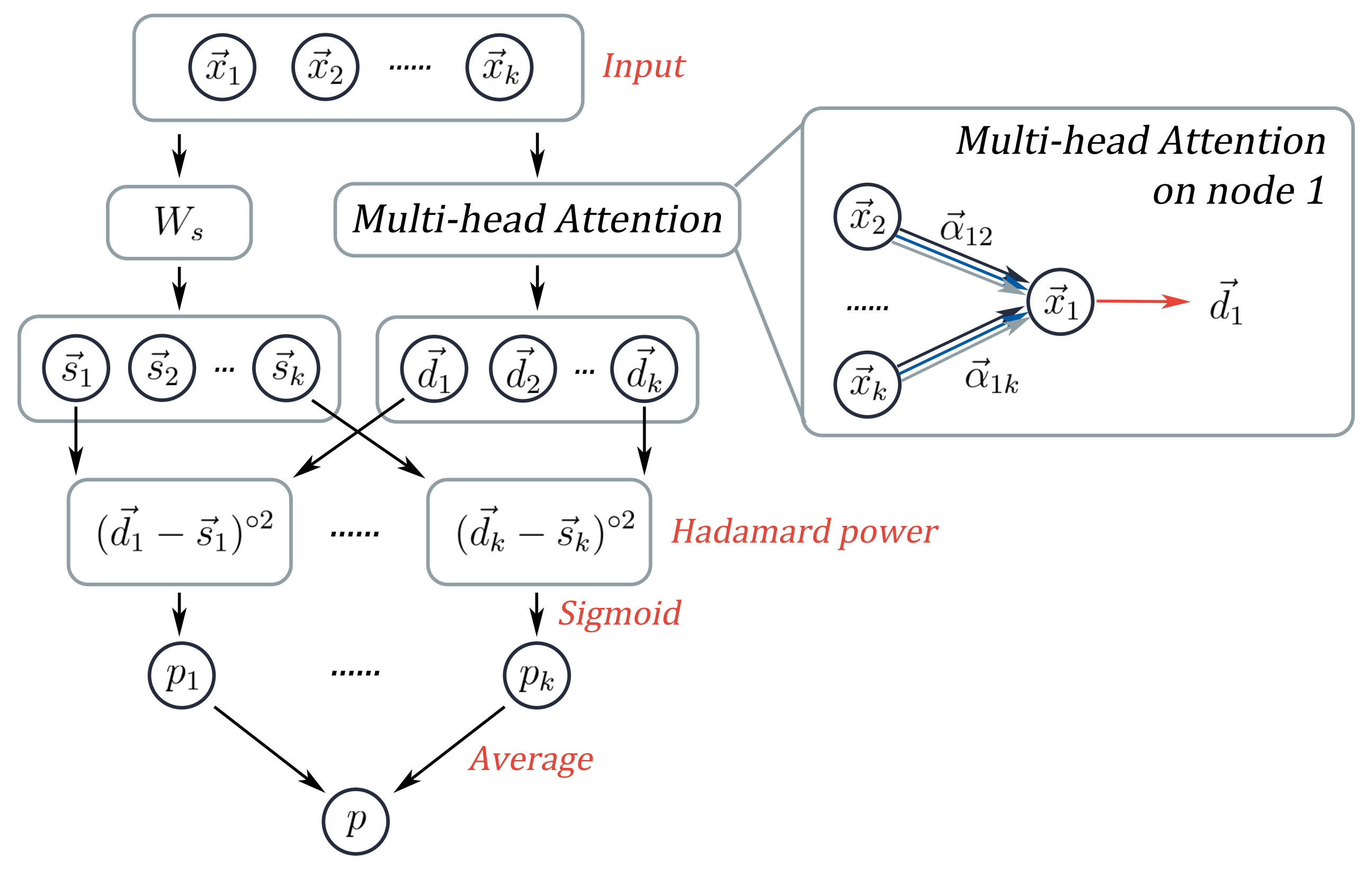}
    \caption{The Hyper-SAGNN architecture from Zhang, et al.~\cite{zhang2020hyper}. The vectors $\vec{x}_1, \vec{x}_2, \hdots, \vec{x}_k$ denote $k$ nodes from a (non-bipartite) hyperedge. We extend this idea to three architectures for bipartite hypergraphs (Figure~\ref{fig:variants}).}
    \label{fig:hypersagnn}
\end{figure}

Currently, the best known model to predict hyperedges is that of Hyper-SAGNN~\cite{zhang2020hyper} (Figure~\ref{fig:hypersagnn}), which uses (1) a self-attention framework to model information flow between nodes of a hyperedge, and (2) a static-vs-dynamic comparative technique to learn hyperedge formation.
But the main reason why it does not deem fit for the bipartite hyperedge scenario is that it captures information flow between \textbf{all} the nodes of a usual (non-bipartite) hyperedge.
And for a per-fixed bipartite hypergraph, the difference is subtler: ``pay heed to the cross-connections, not the self-connections.''
In other words, one need not model the occurrence of individual left and right hyperedges, and instead focus on the connection \textit{between} them.
That is, we need not \textit{predict} left and right hyperedges individually, but only predict the connections between them, since the former is already fixed to be $\F$ and $\F^\prime$.
For the symptoms-diseases case, the reason why Hyper-SAGNN does not apply to the BHP prediction problem is that we are interested in modeling/predicting new diagnoses involving observed symptom-sets and disease-sets.
Paying simultaneous attention to symptoms, diseases and their connections (diagnoses) leads to a learning process that oscillates between the positive and negative classes.
More about this would be discussed in Section~\ref{sec:results:pos_neg_explanation}.

\subsection{Usual Hyper-SAGNN based Self-Attention}

Let us see what effect applying Hyper-SAGNN has on a bipartite hyperedge.
Consider a bipartite hypergraph $\H = (\V, \V^\prime, \F, \F^\prime, \B)$.
Now, let us take two sets $f := \{u_1, u_2, \hdots, u_k\} \subseteq \V$ and $f^\prime := \{u^\prime_1, u^\prime_2, \hdots, u^\prime_{k^\prime}\} \subseteq \V^\prime$ that are ``observed'' to be hyperedges.
Also given are the embeddings of each node: $\x_i, \x^\prime_{i^\prime} \in \mathbb{R}^d$ for nodes $u_i$ and $u^\prime_{i^\prime}$ respectively ($1 \leq i \leq k$, $1 \leq i^\prime \leq k^\prime$).
If $\W_Q, \W_K \in \mathbb{R}^{d \times d_K}$, and $\W_V \in \mathbb{R}^{d \times d_V}$ denote weight matrices for the self-attention framework of Hyper-SAGNN, we have for the potential $(k+k^\prime)$-sized hyperedge $f \cup f^\prime = \{u_1, u_2, \hdots, u_k, u^\prime_1, u^\prime_2, \hdots, u^\prime_{k^\prime}\}$, 
\begin{align}
    a_{ij} = & (\W^T_Q \x_i)^T(\W^T_K \x_j), ~\text{and}~
    a^\prime_{i^\prime j^\prime} = (\W^T_Q \x^\prime_{i^\prime})^T(\W^T_K \x^\prime_{j^\prime}), &\text{ ($\V$ to $\V$ and $\V^\prime$ to $\V^\prime$)} 
    \label{eq:self_attn_a_aprime}\\
    b_{ij^\prime} = & (\W^T_Q \x_i)^T(\W^T_K \x^\prime_{j^\prime}), ~\text{and}~
    c_{i^\prime j} = (\W^T_Q \x^\prime_{i^\prime})^T(\W^T_K \x_{j}) & \text{($\V$ to $\V^\prime$ and $\V^\prime$ to $\V$)} \label{eq:self_attn_b_c}
\end{align}
where $1\leq i, j\leq k$ and $1 \leq i^\prime, j^\prime \leq k^\prime$.
These values are then normalized using the softmax function as follows:
\begin{equation}
\label{eq:self_attn_alpha_beta_gamma}
    \alpha_{ij} =  \frac{\exp(a_{ij})}{z_i},~
    \alpha^\prime_{i^\prime j^\prime} = \frac{\exp(a^\prime_{i^\prime j^\prime})}{z^\prime_{i^\prime}},~
    \beta_{ij^\prime} =  \frac{\exp(b_{i j^\prime})}{z_{i}},~ \gamma_{i^\prime j} = \frac{\exp(c_{i^\prime j})}{z^\prime_{i^\prime}},
\end{equation}
where $z_i := \left( \displaystyle\sum_{t=1}^{k}{\exp(a_{it})} + \sum_{t=1}^{k^\prime}{\exp(b_{it})}\right)$ and $z^\prime_{i^\prime} := \left(\displaystyle\sum_{t=1}^{k^\prime}{\exp(a^\prime_{i^\prime t})} + \sum_{t=1}^{k}{\exp(c_{i^\prime t})}\right)$.

The dynamic embeddings of the hyperedge nodes would then be:
\begin{align}
    \d_i := & \tanh\left(\sum_{l = 1}^{k}{\alpha_{it}\W_V^T\x_t} + \sum_{l = 1}^{k^\prime}{\alpha^\prime_{it}\W_V^T\x^\prime_{t}} \right),\label{eq:self_attn_d}\\
    \d^\prime_{i^\prime} := & \tanh\left(\sum_{t=1}^{k}{\beta_{i^\prime t}\W_V^T\x_t} + \sum_{t=1}^{k^\prime}{\beta^\prime_{i^\prime t}\W_V^T\x^\prime_{t}} \right)\label{eq:self_attn_d_prime}
\end{align}

\subsection{Cross-Attention for Bipartite Hypergraphs}
\label{sec:cat:cat_bih}
Going by the cross-attention paradigm, we introduce three extra attention parameters, amounting to parameters $\W_Q,\W_K,\W^\prime_Q, \W^\prime_K \in \mathbb{R}^{d \times d_K}$ and  $\W_V, \W^\prime_V \in \mathbb{R}^{d \times d_V}$. We redefine $b$, $c$, $\beta$, and $\gamma$ for $1\leq i \leq k$ and $1 \leq i^\prime \leq k^\prime$ as follows:
\begin{align}
    b_{ii^\prime} = &(\W^T_Q \x_i)^T(\W^{\prime T}_K \x^\prime_{i^\prime}),~\text{and}~
    \beta_{ii^\prime} = {\exp(b_{ii^\prime})}\left/{\displaystyle\sum_{t=1}^{k^\prime}{\exp(b_{it})}}\right., \label{eq:cross_attn_b}\\
    c_{i^\prime i} = &(\W^{\prime T}_Q \x^\prime_{i^\prime})^T(\W^T_K \x_{i}),~\text{and}~
    \gamma_{i^\prime i} = {\exp(c_{i^\prime i})}\left/{\displaystyle\sum_{t=1}^{k}{\exp(c_{i^\prime t})}}\right.. \label{eq:cross_attn_c}
\end{align}
Now, the cross-attention dynamic embedding of the left and right hyperedge nodes would then be:
\begin{equation}
    \label{eq:cross_attn_delta}
    \deltabf_i := \tanh\left(\sum_{t=1}^{k^\prime}{\beta_{it}\W_V^{\prime T}\x^\prime_t}\right),
    \deltabf^\prime_{i^\prime} := \tanh\left(\sum_{t=1}^{k}{\gamma_{i^\prime t}\W_V^T\x_t}\right).
\end{equation}
Finally, we have new embeddings $\deltabf_1, \deltabf_2, \hdots, \deltabf_k$ of left nodes and $\deltabf^\prime_1, \deltabf^\prime_2, \hdots, \deltabf^\prime_{k^\prime}$ of right nodes, all $d_V$-dimensional vectors.

%% file: sections/catsetmat.tex
\begin{figure}
    \centering
    \begin{minipage}{0.33\linewidth}
    \subfloat[\algocsmx]{\centerline{\includegraphics[width=0.9\linewidth, page=1]{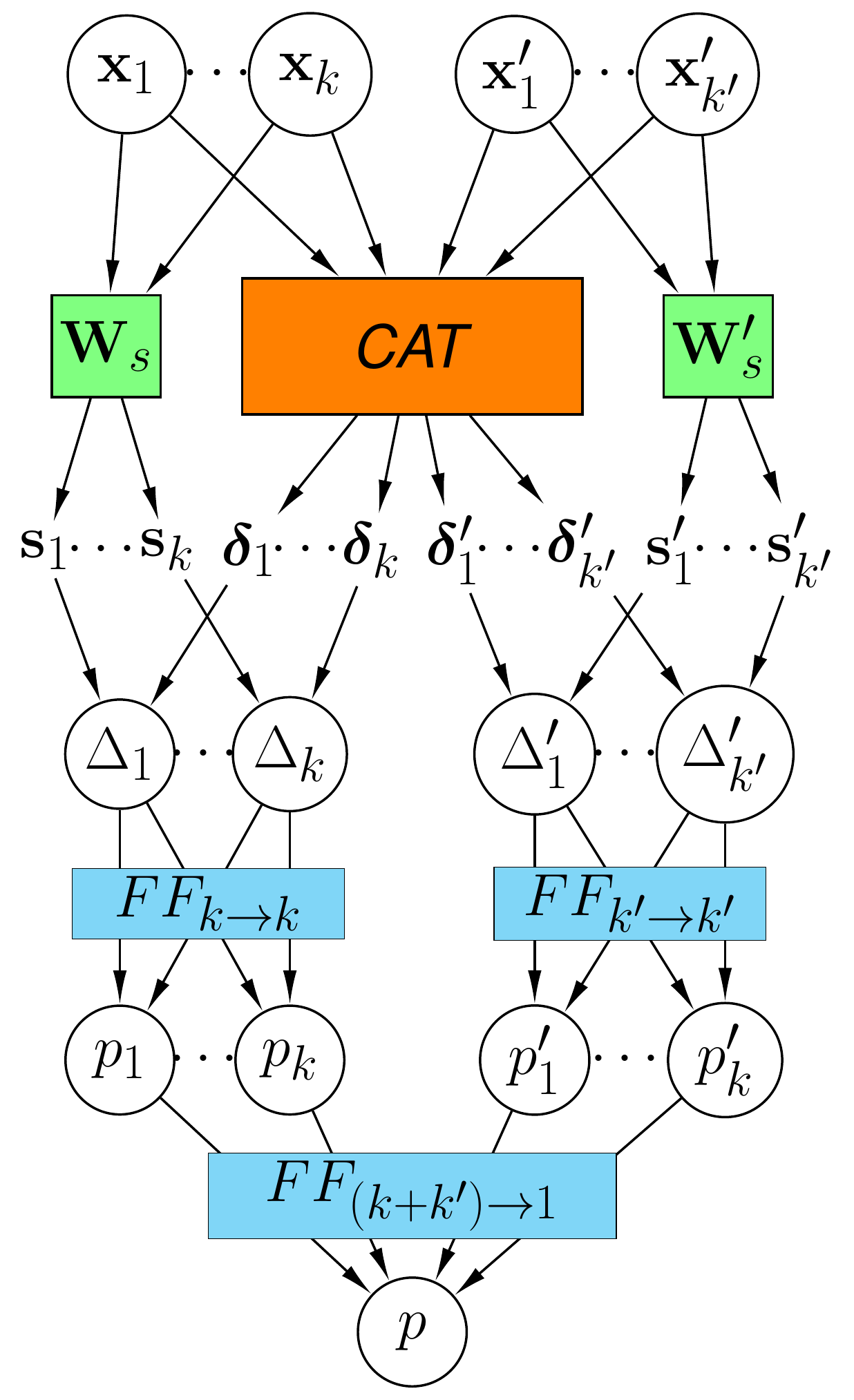}}}
\end{minipage}%
\begin{minipage}{0.33\linewidth}
    \subfloat[\algocsmsx]{\centerline{\includegraphics[width=0.9\linewidth, page=2]{figures/catsetmat.pdf}}}
\end{minipage}%
\begin{minipage}{0.33\linewidth}
    \subfloat[\algocsmsxs]{\centerline{\includegraphics[width=0.9\linewidth, page=3]{figures/catsetmat.pdf}}}
\end{minipage}%
    \caption{Neural network architectures of the three variants of the \algocsm{} algorithm that we propose. (a) Only one cross-attention layer (\texttt{X}). (b) An additional pair of self-attention layers (\texttt{S}) before cross-attention (\texttt{SX}). (c) Another additional pair of self-attention layers after cross-attention (\texttt{SXS}).}
    \label{fig:variants}
\end{figure}

Akin to Hyper-SAGNN (Figure~\ref{fig:hypersagnn}), we define a neural network architecture called \catsetmat{} (\textit{C}ross \textit{AT}tention for \textit{SET} \textit{MAT}ching) that has both self- as well as cross-attention layers.
\catsetmat{} uses structures defined in Section~\ref{sec:cat:cat_bih} that are derived from the self-attention layers in Hyper-SAGNN.
A detailed network architecture is depicted in Figure~\ref{fig:variants}, with three of its variants: one with no self-attention, and the others with one and two extra pairs of the same.
And as explained before, we use the architecture to solve the set matching prediction (SMP) problem which in turn solves the bipartite hyperedge prediction (BHP) problem -- a problem we had termed bipartite hyperedge set matching prediction (BHSMP).

Basically, it commences with the vector representations of two sets of hyperedges -- one left ($\{\x_1, \x_2, \hdots, \x_k\}$), and the other right ($\{\x^\prime_1, \x^\prime_2, \hdots, \x^\prime_{k^\prime}\}$).
These inputs are fed into two Hyper-SAGNN-like self-attention (SAT) blocks \sat{} and \satp{} separately, which have their respective parameter sets $\W^{(\sat)}_Q, \W^{(\sat)}_K, \W^{(\sat)}_V$ and $\W^{(\satp)}_Q, \W^{(\satp)}_K, \W^{(\satp)}_V$).
The \sat-blocks give out revised embeddings $\y_1, \y_2, \hdots, \y_k$ and $\y^\prime_1, \y^\prime_2, \hdots, \y^\prime_{k^\prime}$.
\textbf{Although this applies one round of attention to each of the left and right hyperedges, no information about the cross-connections between them have been learned by our model.}

Here is where the cross-attention (CAT) block -- \cat{} -- comes into picture, not-to-mention, with its own set of parameters $\W^{(\cat)}_Q$, $\W^{(\cat)}_K$, $\W^{(\cat)}_V$, $\W^{\prime(\cat)}_Q$, $\W^{\prime(\cat)}_K$, $\W^{\prime(\cat)}_V$.
The \cat{} block takes the revised embeddings through equations~\ref{eq:cross_attn_b}--\ref{eq:cross_attn_delta} and returns fresh embedding vectors $\deltabf_1, \deltabf_2, \hdots, \deltabf_k$ and $\deltabf^\prime_1, \deltabf^\prime_2, \hdots, \deltabf^\prime_{k^\prime}$ -- vectors we call \textit{dynamic embeddings}, just as Zhang, et al.~\cite{zhang2020hyper} do for Hyper-SAGNN.
In parallel is a layer of static-weights $\W_s, \W^\prime_s \in \mathbb{R}^{d \times d_s}$ that simply transforms the original left vectors $\x_i$ and right vectors $\x^\prime_{i^\prime}$ into \textit{static embeddings} $\s_1, \s_2, \hdots, \s_k$ and $\s^\prime_1, \s^\prime_2, \hdots, \s^\prime_{k^\prime}$ defined by $\s_i := \tanh\left(\W_s^T\x_i\right)$ and $\s^\prime_{i^\prime} := \tanh\left(\W^{\prime T}_{s}\x^\prime_{i^\prime}\right)$ respectively, where $1 \leq i \leq k$ and $1 \leq i^\prime \leq k^\prime$.
Following Zhang et al.~\cite{zhang2020hyper}, we compare the static and dynamic embeddings using the Hadamard square operator, resulting in vectors $\Delta_1, \Delta_2, \hdots, \Delta_k$ and $\Delta^\prime_1, \Delta^\prime_2, \hdots, \Delta^\prime_{k^\prime}$ defined as $\Delta_i := (\deltabf_i - \s_i)^{\circ2}$ and $\Delta^\prime_{i^\prime} := (\deltabf^\prime_{i^\prime} - \s^\prime_{i^\prime})^{\circ2}$.
Finally, two positional feed-forward layers~\cite{zhang2020hyper} $FF_{k \rightarrow k}$ and $FF_{k^\prime \rightarrow k^\prime}$ followed by a consolidating layer $FF_{(k+k^\prime \rightarrow 1)}$ computes the probability $p$ that the set of nodes match or not.


\subsection{Variants of \algocsm{}}
We use altogether three variants of \algocsm{}: \algocsmx{} (only \cat{}), \algocsmsx{} (a pair of \sat{} followed by a \cat{}), and \algocsmsxs{} (\algocsmsx{} with another pair of \sat{} modules following \cat{}).
Architecture diagrams for all the three variants have been shown in Figure~\ref{fig:variants}.

%% file: sections/relwork.tex
As far as usual hypergraphs are concerned, they have not been studied as much as graphs (some earlier works are are~\cite{zhou2007learning,li2013link,bonacich2004hyper}).
Of some recent approaches~\cite{benson2018simplicial,tu2018structural,zhang2018beyond,bai2019hypergraph,feng2019hypergraph,huang2019hyper2vec,jiang2019dynamic,payne2019deep,yadati2019link,yadati2019hypergcn,bandyopadhyay2020line,zhang2020hyper} for hyperedge prediction and hypergraph embedding, the most recent one, \textit{viz.}, Hyper-SAGNN~\cite{zhang2020hyper} performs the best.
It uses a self-attention based model wherein each hyperedge is handled separately.
Our approach is strongly based on theirs, and the whole concept of cross-attention (and a combination thereof with self-attention) extends their framework to set matching as well.

It is to be noted that left and right hyperedges in a bipartite hypergraph are basically sets of nodes; hence machine learning techniques to handle sets become relevant here.
Deep set embedding calls for permutation-invariant neural networks, and there has been a considerable amount of work on this topic~\cite{vinyals2015order,zaheer2017deep,wagstaff2019limitations,lee2019set,zhang2019fspool,meng2019hats,wendler2019powerset}, of which we use  FSPool~\cite{zhang2019fspool}, a sort-pooling based technique, as one of our baselines.
Moreover, deep set-to-set matching has been performed on image data~\cite{saito2019deep}, but it does not apply to our problem since it uses a single universal set, as opposed to two disjoint ones in the case of a bipartite hypergraph.

Bipartite networks are so essential that Guillaume et al.~\cite{guillaume2004bipartite} have argued for an underlying bipartite structure in \textit{all} networks.
Of late, neural network techniques for usual graphs have become widely recognized~\cite{grover2016node2vec,kipf2016semi,hamilton2017inductive,velivckovic2017graph} (ref.~Wu et al.~\cite{wu2020comprehensive} for a survey).
But the same is not true for deep bipartite networks.
The \catsetmat{} architecture we propose inherits attention from GAT~\cite{velivckovic2017graph} and weakly relates to BGNN~\cite{he2019bipartite}.
\textbf{Furthermore, to the best of our knowledge, there is no work that explores a bipartite hypergraph from the perspective of network science}, and all of them~\cite{zverovich2006bipartite,annamalai2016finding,heismann2014hypergraph} belong to the domain of discrete mathematics.

\begin{table*}[!b]
    \caption{The list of bipartite hypergraph datasets used in this work, along with their vital statistics: \# left nodes $|\V|$, \# right nodes $|\V^\prime|$, \# left hyperedges $|\F|$, \# right hyperedges $|\F^\prime|$, and \# bipartite hyperedges $|\B|$.}
    \label{tab:datasets}
    \centering
    \input{tables/datasets.tex}
\end{table*}

\begin{figure}[!b]
    \centering
    \begin{minipage}{0.33\linewidth}
    \subfloat[mag-acm-ak]{\centerline{\includegraphics[width=0.9\linewidth, page=1]{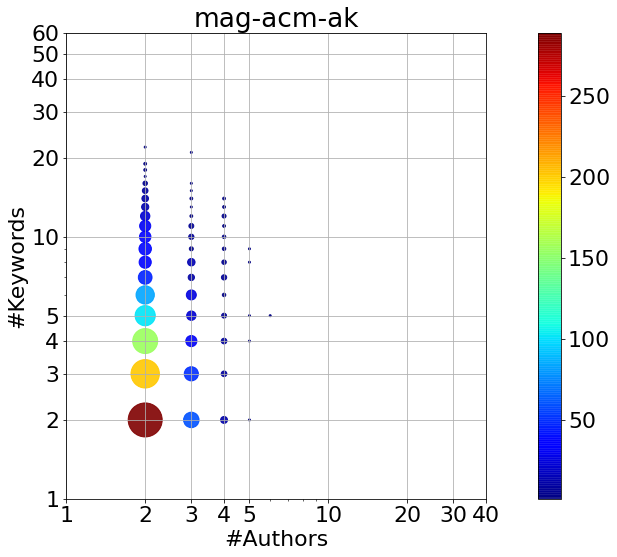}}}
\end{minipage}%
\begin{minipage}{0.33\linewidth}
    \subfloat[tmdb-cc]{\centerline{\includegraphics[width=0.9\linewidth, page=1]{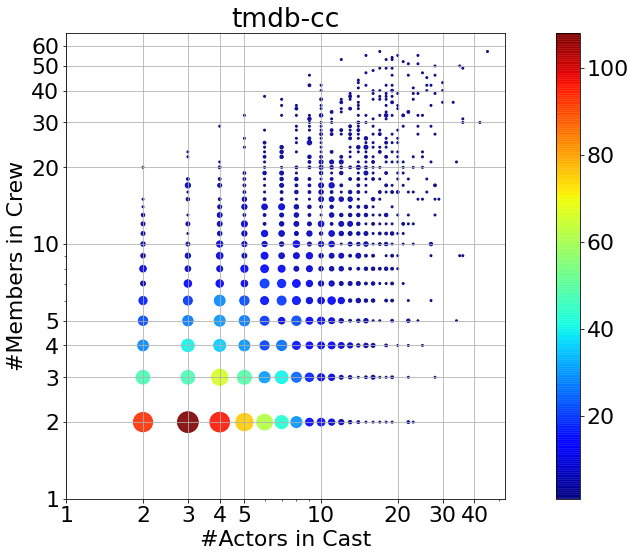}}}
\end{minipage}%
\begin{minipage}{0.33\linewidth}
    \subfloat[tmdb-ck]{\centerline{\includegraphics[width=0.9\linewidth, page=1]{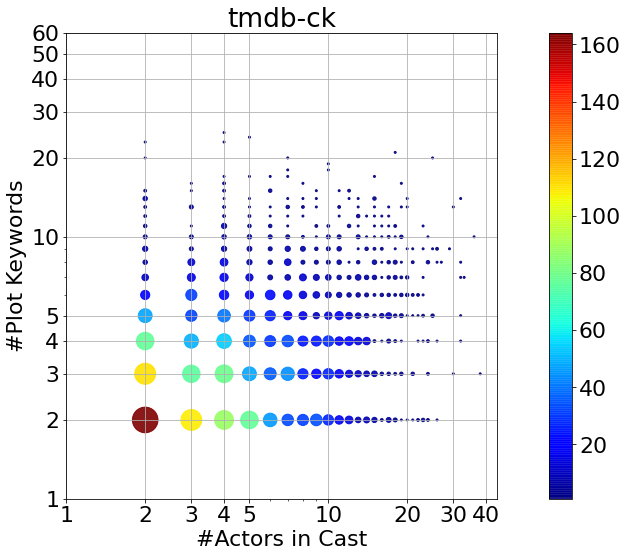}}}
\end{minipage}%
    \caption{Size distributions of the left and right hyperedges in the bipartite hypergraph for the threedatasets. On X-axis is the size of the left hyperedges, and on Y-axis, that of right hyperedges.}
    \label{fig:dataset_dist}
\end{figure}

%% file: tables/datasets.tex
\begin{tabular}{llllrrrrr}
\toprule
\textbf{Dataset} & Left nodes ($\V$) & Right nodes ($\V^\prime$) & $|\V|$ & $|\V^\prime|$ & $|\B|$ & $|\F|$ & $|\F^\prime|$\\
\midrule
\dataicc   & Cast (actors) & Crew (other members) & 4,556 &        3,802 & 2,825 &     2,824 &            2,744 \\
\dataick   & Cast (actors) & Plot keywords & 3,156 &        1,256 & 2,669 &     2,656 &            2,621 \\
\datamacm  & Authors & Keywords & 1,059 &        2,338 & 1,388 &       847 &            1,379 \\
\bottomrule
\end{tabular}                                                                                                                                            

%% file: sections/exp.tex
We perform bipartite hyperedge prediction experiments on some real-world datasets.
Apart from the baselines described in Section~\ref{sec:exp:baselines}, we use three versions of \algocsm{}: \algocsmx{} (only \cat{}), \algocsmsx{} (a pair of \sat{} followed by a \cat{}), and \algocsmsxs{} (\algocsmsx{} with another pair of \sat{} modules following \cat{}).
Refer Table~\ref{tab:datasets} for details about datasets and Figure~\ref{fig:dataset_dist} for distribution of bipartite hyperedge sizes therein.
Details about data preparation and other reproducibility details could be found in Appendix~\ref{app:exp}.

\subsection{Baselines}
\label{sec:exp:baselines}
As discussed in Section~\ref{sec:relwork}, we prepare four classes of baselines: node2vec-based, bipartite-graph-based, set-embedding-based and hyperedge-prediction-based, each of which has been described below.
\begin{enumerate}
\item \textbf{node2vec-based}:\\ \algoncme{}, \algonfme{}, \algoncmi{}, \algonfmi{}.
More details could be found in Appendix~\ref{app:exp:baselines:n2v}.
 \item \textbf{Bipartite-graph-based}: Since our hypergraph could be interpreted as a bipartite ``graph'' (Observation~\ref{obs:hyg_as_graph}), we prepare some baselines on bipartite versions of two popular link prediction algorithms~\cite{liben2007link}: \textit{Common Neighbor (CN)}~\cite{newman2003structure} and \textit{Adamic Adar (AA)}~\cite{adamic2003friends}.
 Please refer to Appendix~\ref{app:exp:baselines:bip} for more details on how these features are computed for a bipartite hypergraph.
 \item \textbf{Set-embedding-based} (\textbf{FSPool}): For this, we first convert the node embeddings obtained from node2vec~\cite{grover2016node2vec} (with dimension $16$, since it showed best results) of each of the left and right hyperedges into set-embeddings using FSPool~\cite{zhang2019fspool}, and then learn a simple classifier.
 \item \textbf{Hyperedge-prediction-based} (\textbf{Hyper-SAGNN}): We use the existing best performing algorithm Hyper-SAGNN~\cite{zhang2020hyper} for hyperedge prediction, wherein we interpret our bipartite hypergraph as a usual one as per Observation~\ref{obs:hyg_as_graph}. We use the same experiment settings as Zhang et al.~\cite{zhang2020hyper} and their architecture as shown in Figure~\ref{fig:hypersagnn}.
\end{enumerate}

%% file: sections/results.tex
\begin{table*}[!t]
    \caption{Results (\%AUC) for the bipartite hyperedge prediction problem}
    \label{tab:results}
    \centering
    \input{tables/results.tex}
\end{table*}

The results for hyperedge prediction on a few datasets have been listed in Table~\ref{tab:results}, where each baseline class~(see Section~\ref{sec:exp:baselines}) has been separated by a horizontal line.
For each algorithm, and each dataset, we report the \%AUC test scores for the SMBHP problem (see Definition~\ref{def:smbhp}).
In each case, we see that our algorithm \catsetmat{} performs the best unanimously.
More observations are described in the following sections (Sections~\ref{sec:results:baselines}--\ref{sec:results:catsetmat}).
An explanation of the poor performance of some state-of-the-art algorithms has been provided in Section~\ref{sec:results:pos_neg_explanation}.

\subsection{Baselines}
\label{sec:results:baselines}
Although bipartite link prediction algorithms (\algolpminaamean--\algolpavgcnmean) perform poorly with AUCs ranging from $43$--$55\%$ and $37$--$50\%$ for the first two datasets, for \datamacm{}, some of them perform considerably better with AUCs going up to $64.5\%$ as well.
These algorithms heavily depend on the interconnectivity between the left and right hyperedges, and it seems to be the best in the \datamacm{} dataset.
The next set of algorithms -- the four node2vec-based ones -- perform much better than the foregoing group, but still lag behind \catsetmat{} by huge margins. Surprisingly, the bipartite graph based algorithms outperform the node2vec ones for the last dataset. This only shows that performances vary dramatically from dataset to dataset.

\begin{figure*}[htb]
    \centering
\begin{minipage}{0.33\linewidth}
    \subfloat[tmdb-cc]{\centerline{\includegraphics[width=1.1\linewidth, page=1]{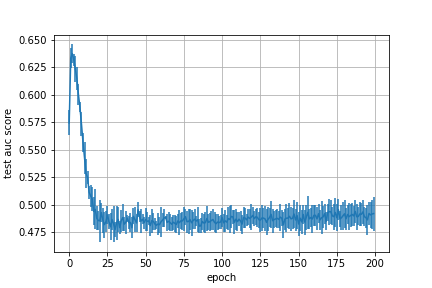}}}
\end{minipage}%
\begin{minipage}{0.33\linewidth}
    \subfloat[tmdb-ck]{\centerline{\includegraphics[width=1.1\linewidth, page=1]{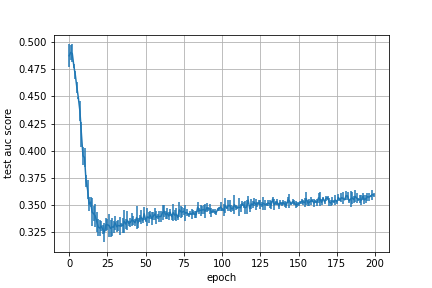}}}
\end{minipage}%
\begin{minipage}{0.33\linewidth}
    \subfloat[mag-acm-ak]{\centerline{\includegraphics[width=1.1\linewidth, page=1]{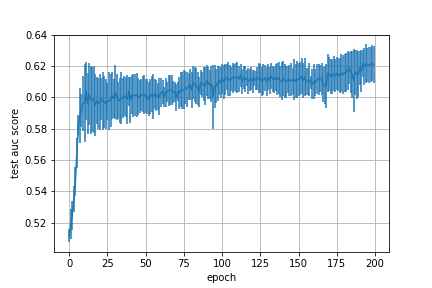}}}
\end{minipage}%
    \caption{AUC learning curves for \algofsp{} on all datasets.}
    \label{fig:fspool_results}
\end{figure*}

\begin{figure*}[htb]
    \centering
\begin{minipage}{0.33\linewidth}
    \subfloat[tmdb-cc]{\centerline{\includegraphics[width=1.1\linewidth, page=1]{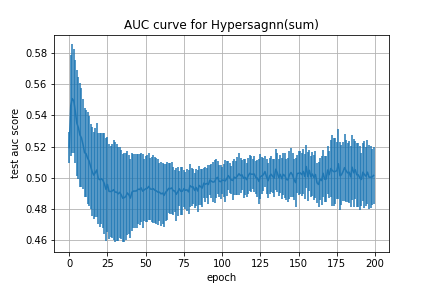}}}
\end{minipage}%
\begin{minipage}{0.33\linewidth}
    \subfloat[tmdb-ck]{\centerline{\includegraphics[width=1.1\linewidth, page=1]{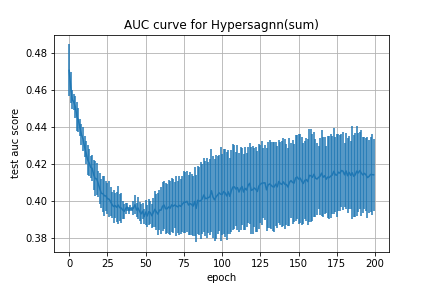}}}
\end{minipage}%
\begin{minipage}{0.33\linewidth}
    \subfloat[mag-acm-ak]{\centerline{\includegraphics[width=1.1\linewidth, page=1]{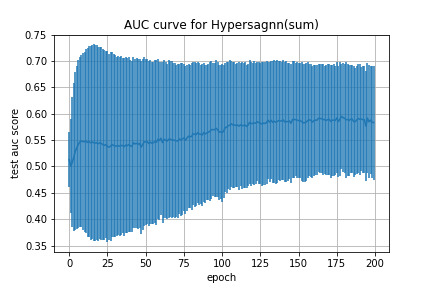}}}
\end{minipage}%
    \caption{AUC learning curves for \algohsnn{} (the ``sum'' variant from~\cite{zhang2020hyper}) on each dataset.}
    \label{fig:hypersagnn_results}
\end{figure*}

The next two algorithms -- \algofsp{} and \algohsnn{}, though being cutting-edge deep learning approaches in their respective fields (\textit{viz.}, set and hyperedge embeddings respectively), fail to capture the desired bipartite hypergraph structure.
The main reason for this, as has been explained in Section~\ref{sec:results:pos_neg_explanation}, is that they struggle to learn a consistent representation of nodes since the same set of nodes get involved in bipartite hyperedges (positive class) as well as bipartite non-hyperedges (negative class) simultaneously.
Moreover, \algohsnn{} also shows quite a high amount of deviation ($13\%$; as per Table~\ref{tab:results}), which is undesirable of any machine learning algorithm.
But this only shows the huge dependence of \algohsnn{} on the data preparation process (\textit{i.e.}, negative-sampling and train-test split) that we perform five times. More about this has been discussed in Section~\ref{sec:results:pos_neg_explanation}, where we use the learning curves (Figures~\ref{fig:fspool_results} and \ref{fig:hypersagnn_results}) to compare \algofsp{}'s and \algohsnn{}'s respective behaviour \textit{w.r.t.} \algocsm{}.

\subsection{CATSETMAT}
\label{sec:results:catsetmat}
\begin{figure}[htb]
    \centering

\begin{minipage}{0.33\linewidth}
    \subfloat[tmdb-cc]{\centerline{\includegraphics[width=1.1\linewidth, page=1]{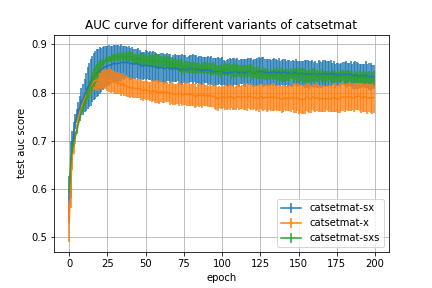}}}
\end{minipage}%
\begin{minipage}{0.33\linewidth}
    \subfloat[tmdb-ck]{\centerline{\includegraphics[width=1.1\linewidth, page=1]{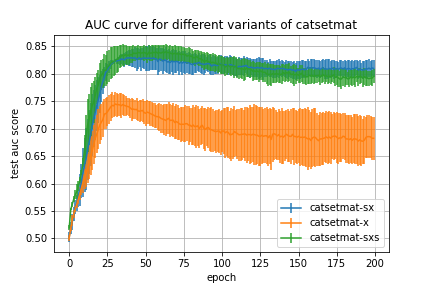}}}
\end{minipage}%
\begin{minipage}{0.33\linewidth}
    \subfloat[mag-acm-ak]{\centerline{\includegraphics[width=1.1\linewidth, page=1]{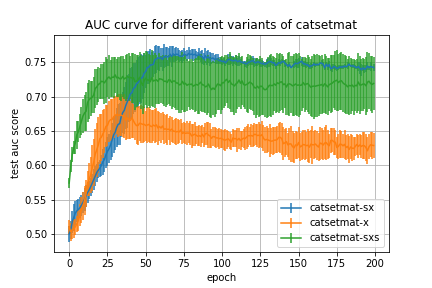}}}
\end{minipage}%
    \caption{Comparison of AUC learning curves for all variants of \algocsm{} on all datasets.}
    \label{fig:catsetmat_results}
\end{figure}

Finally, the last three algorithms -- the ones we have proposed as part of this work -- are the best performing ones among the lot. For \dataicc{}, we see that our best method is $13.56\%$ better than the group-wise next best (\textit{i.e.}, \algonfme), which has an AUC of only $77.68\%$.
Another of our methods (\algocsmsxs) performs the best for \dataick, with a much higher improvement of $33.06\%$ as compared to \algoncme, the group-wise next best.
For both these datasets, we obtain quite high AUC values of around $84$--$88\%$, but the same is not true for \datamacm{}, for which the best performance is limited to $76.62\%$, albeit given by our algorithm \algocsmsx.
Although this is a much lower score as compared to the other two datasets, but it is $18.65\%$ higher than \algolpavgaamean{}, which is the best performing baseline.
Barring our algorithms \algocsmx, \algocsmsx, and \algocsmsxs, no other algorithm consistently performs well on all the three datasets.
While at least one dataset deemed problematic for each of the baselines, these three algorithms remained consistent for all datasets.
Moreover, no baseline touched the $80\%$ AUC mark except the \algocsm{}-based algorithms.
We credit this success to the \textit{cross-attention paradigm}, which focuses on the cross-links more than on the left and right hyperedges, thereby avoiding the positive-negative dilemma as explained in Section~\ref{sec:results:pos_neg_explanation}, where we make use of Figure~\ref{fig:catsetmat_results} to explain the better performance of our algorithm.

Although the \cat{} module (see Figure~\ref{fig:variants}) is our main contribution in this work, adding more \sat{} layers is expected to model the individual left- and right-hyperedges as well, since it might capture some inner domain-specific relational structure in each of them.
We can clearly see this effect being illustrated in Table~\ref{tab:results}, wherein adding only one pair of \sat{} modules before \cat{} (in \algocsmsx) boosts the performances of a purely \cat-based algorithm \algocsmx{} by upto $3.92\%$, $12.59\%$, and $13.27\%$ for the three datasets respectively.
While the addition of another pair of \sat{} modules (\algocsmsxs) shows the potential of improving the AUC performance by an additional $1.93\%$ for \dataicc, but same is not true for the \dataick{} and \datamacm{}, where the performance instead drops by $2.57\%$ and $1.31\%$ respectively.
This effect could be attributed to over-fitting due to an increase in model parameters.
As a result of this, we conclude that addition of more number of \sat{} modules would not further boost the performance, and so, it should be limited to a single pair of \sat{} modules before \cat{}, as done in \algocsmsx.
The difference is also clear from the learning curves in Figure~\ref{fig:catsetmat_results}, where the relative epoch-by-epoch performance of each variant could be observed.

\subsection{The Positive-Negative Dilemma}
\label{sec:results:pos_neg_explanation}
It is clear that our algorithm works well, and hence the baselines fail to capture the ``bipartiteness'' for
hyperlink prediction.
Figures~\ref{fig:fspool_results},~\ref{fig:hypersagnn_results}, and~\ref{fig:catsetmat_results} depict the AUC learning curves for \algofsp{}, \algohsnn{}, and \algocsm{} respectively on the three datasets.
We could see that unlike \algocsm{}, the baselines \algofsp{} and \algohsnn{} do not seem to converge, and keep oscillating around an AUC of 50\%.

We reason for this behavior as follows. 
A careful look at the definition of a per-fixed bipartite hypergraph (Definition~\ref{def:bipartite_hypergraph}) would reveal a straightforward yet important fact: \textit{that left and right sets of hyperedges are fixed in advance}.
What makes this factor interesting is another definition: that of bipartite non-hyperedges (again, Definition~\ref{def:bipartite_hypergraph}) or the negative class, which ensures that \textit{each negative sample would be formed from the same fixed left and right hyperedges that formed samples from the positive class}.
What is striking to observe is that there is nothing that stops a left hyperedge $f$ that involves in forming a positive sample from getting involved in forming a negative sample as well.
The same could also be said for a right hyperedge $f^\prime$.
As a result, we observe that the positive and the negative samples are built from mostly (if not entirely) the same left and right hyperedges.

When an algorithm that treats the entire hyperedge $f \cup f^\prime$ as a whole is deployed to embed it, it keeps getting confused, as to how to train the parameters to account for both classes simultaneously.
Thus, it does not settle for a stable output for both $f$ as well as $f^\prime$, owing to $f$ and $f^\prime$'s simultaneous association with both the classes.
This is what algorithms like \algohsnn{} and \algofsp{} suffer from, when it comes to a per-fixed bipartite hypergraph.

\subsection{An Additional Experiment to Understand the Positive-Negative Dilemma}
\label{sec:results:pos_neg_experiment}
The attention parameters $\W_Q$, $\W_K$, and $\W_V$ that the Hyper-SAGNN formulation uses keep getting modified as per the positive and negative class labels, and since in our case, the bipartite hyperedges and bipartite non-hyperedges both use the same set of left and right hyperedges (owing to them being fixed), we perform an experiment that undoes this effect.
For this, we change our negative sampling method to a ``sized-random''~\cite{patil2020negative} technique and use the following, we essentially change the ``per-fixed'' hypergraph (from Definition~\ref{def:bipartite_hypergraph}) to a ``simple''
hypergraph.

The exact procedure is as follows:
Given a hypergraph $\H = (\V, \V^\prime, \F, \F^\prime, \B)$:

\begin{figure*}
    \centering
\begin{minipage}{0.5\linewidth}
    \subfloat[\datamacm{} (non-per-fixed)]{\centerline{\includegraphics[width=1.1\linewidth, page=1]{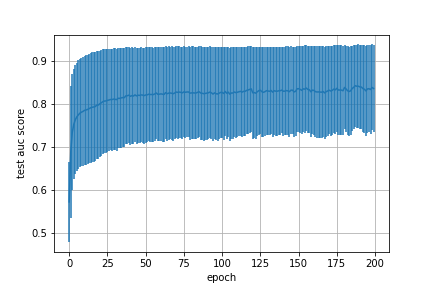}}}
\end{minipage}%
\begin{minipage}{0.5\linewidth}
    \subfloat[\dataicc{} (non-per-fixed)]{\centerline{\includegraphics[width=1.1\linewidth, page=1]{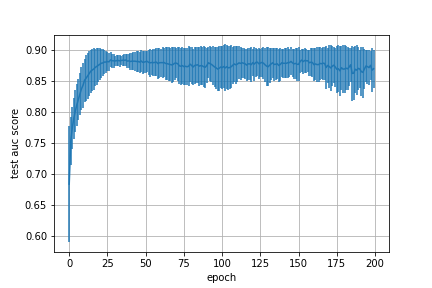}}}
\end{minipage}%
    \caption{AUC learning curves for \algohsnn{} with sized-random negative sampling~\cite{patil2020negative} on non-per-fixed variants of two of our datasets (refer to Section~\ref{sec:results:pos_neg_experiment} for the exact procedure).}
    \label{fig:randomsampling}
\end{figure*}

\begin{enumerate}
    \item Initialize $\hat{\B} = \{\}$, the set of bipartite non-hyperedges.
    \item Pick a hyperedge $b = f \cup f^\prime \in \B$
    \item Pick a random set $\hat{f} \subseteq \V$ of left nodes such that $|\hat{f}| = |f|$, and a random set $\hat{f}^\prime \subseteq \V$ of right nodes such that $|\hat{f}^\prime| = |f^\prime|$. Define $\hat{b} := \hat{f} \cup \hat{f}^\prime$.
    \item If $\hat{b} \in \B \cup \hat{\B}$, repeat Step 3. Else, add the new bipartite non-hyperedge $b$ to $\hat{\B}$.
    \item Repeat steps 2--4 until $|\hat{\B}| = 5\cdot|\B|$.
\end{enumerate}

Performing \algohsnn{} on this simpler hypergraph, \textbf{surprisingly enough, makes the result totally different!}
\algohsnn{} now reports a test AUC of a whopping $98\%$ (see Figure~\ref{fig:randomsampling}) on \datamacm{}, a dataset that was the toughest to handle as per Table~\ref{tab:results}.
It is also advisable to see how well the learning curve boosts-up with this version of the data.
It is clear from the foregoing arguments that there exists a positive-negative dilemma in usual hyperedge embedding approaches as far as bipartite hypergraphs are concerned.
The unusually higher variance of $13\%$ observed in Table~\ref{tab:results} for \algohsnn{} on \datamacm{} could be easily attributed to this dilemma.

%% file: tables/results.tex


\begin{tabular}{lccc}
\toprule
        Algorithm &            \dataicc &            \dataick &            \datamacm \\
\midrule
 \algolpminaamean &  $43.9987\pm0.9787$ &  $37.6217\pm1.2686$ &   $49.9388\pm1.1913$ \\
 \algolpmincnmean &  $43.8015\pm0.9613$ &  $37.5078\pm1.2589$ &   $49.8958\pm1.1743$ \\
 \algolpmaxaamean &  $53.0463\pm0.4581$ &  $49.4547\pm0.5887$ &   $63.5835\pm0.5314$ \\
 \algolpmaxcnmean &  $52.5955\pm0.4971$ &  $49.3627\pm0.5697$ &   $62.0365\pm0.6099$ \\
 \algolpavgaamean &  $54.6746\pm0.9042$ &  $46.9037\pm0.7630$ &   $\mathit{64.5778\pm1.6425}$ \\
 \algolpavgcnmean &  $54.0634\pm0.9388$ &  $46.7163\pm0.7498$ &   $63.3550\pm1.7340$ \\
 \midrule
        \algoncme &  $77.4467\pm1.1838$ &  $\mathit{63.5867\pm0.4989}$ &   $59.2567\pm2.0266$ \\
        \algonfme &  $\mathit{77.6833\pm1.5083}$ &  $62.8900\pm0.1606$ &   $59.0933\pm2.2257$ \\
        \algoncmi &  $52.4433\pm1.5320$ &  $51.9333\pm1.2777$ &   $32.8867\pm1.1585$ \\
        \algonfmi &  $53.9433\pm1.0404$ &  $52.8267\pm1.1521$ &   $31.7933\pm1.6376$ \\
\midrule
         \algofsp &  $63.8305\pm0.8315$ &  $50.2596\pm1.3653$ &   $63.4810\pm1.6438$ \\
\midrule
     \algohsnn &  $55.6071\pm2.9784$ &  $47.0969\pm1.3497$ &  $63.0752\pm13.0197$ \\
\midrule
        \algocsmx &  $83.2761\pm2.2315$ &  $74.6650\pm2.1525$ &   $67.6457\pm3.0129$ \\
       \algocsmsx &  $86.5461\pm3.3937$ &  $\mathbf{84.0697\pm0.9814}$ &   $\mathbf{76.6221\pm0.9676}$ \\
      \algocsmsxs &  $\mathbf{88.2177\pm0.8268}$ &  $81.9037\pm2.3348$ &   $75.6302\pm3.0833$ \\
\bottomrule
\end{tabular}

%% file: sections/conc.tex
A bipartite hypergraph is a peculiar data structure almost never studied in network analysis.
We formalize almost all notions of this structure with standard notations used for graphs and usual hypergraphs, and introduce the bipartite hyperedge prediction problem.
However, since we were able to establish an equivalence between this problem and set-matching, we could propose a solution for the latter and use it for the former.
We could successfully posit that focusing on cross-attention plays a significant role in capturing bipartite relations well.
An important insight that we were able establish was the existence of a positive-negative dilemma in existing cutting-edge algorithms like Hyper-SAGNN designed for the very job of hyperedge prediction.

The behavior of bipartite hypergraphs is close to unknown among researchers.
We try to shed some light on one aspect of their application on a problem having real-world implications -- the hyperedge prediction problem. But a lot of problems remain unanswered.
Also, a variety of higher-order relations still remain to be modeled: \textit{e.g.}, a $k$-partite hypergraph. Another huge area that needs attention is the preparation and analysis of more and more bipartite hypergraph real-world datasets. We leave these problems as future work.

%% file: sections/supplementary_content.tex
\section{Proofs}
\label{app:proofs}

\subsection{Proof of Lemma 1}
\label{app:proofs:lemma1}
\begin{proof}
Given a bipartite hypergraph $\H = (\V, \V^\prime, \F, \F^\prime, \mathcal{B})$, consider $\F$ and $\F^\prime$ as two sets-of-sets. Now consider the following claim:
\vspace{-0.2cm}
\begin{myquote}{0.3cm}
\textbf{Claim}: \textit{Sets $\mathbb{B}(\F, \F^\prime)$ (see Definition 1) and $\F \times \F^\prime$ are equivalent.}
\vspace{-0.4cm}
\begin{proof}[Proof of Claim]
The map $\sigma: \mathbb{B}(\F, \F^\prime) \rightarrow \F \times \F^\prime$ defined by $x \mapsto (x \cap \V, x \cap \V^\prime)$ is bijective since its inverse $\sigma^{-1}$ is defined by $(x, y) \mapsto x \cup y$. Hence, the claim.
\end{proof}
\end{myquote}
\vspace{-0.2cm}
\noindent Using the same bijective map in the \textit{Proof of Claim} above, we have:
\begin{equation*}
    \forall b \in \B,~\exists f := b\cap\V \in \F \text{ and }~\exists f^\prime := b\cap\V^\prime \in \F^\prime
\end{equation*}
Hence the relation:
\begin{equation*}
\label{eq:mh}
    \M(\H) := \{(b \cap \V, b \cap \V^\prime) \mid b \in \B\} \subseteq \F \times \F^\prime
\end{equation*}
is a set matching over sets-of-sets $\F$ and $\F^\prime$ as per Definition 3.
\end{proof}

\subsection{Proof of Lemma 2}
\label{app:proofs:lemma2}
\begin{proof}
Given $\H = (\V, \V^\prime, \F, \F^\prime, \B)$, fix $\P\subseteq\B$ and $\N\subseteq\hat{\B}$ as positive and negative classes respectively.
Consider hyperedges $f, \hat{f} \in \F$, $f^\prime, \hat{f}^\prime \in \F^\prime$ be such that $f\cup f^\prime \in \P$ and $\hat{f}\cup \hat{f}^\prime \in \N$.

Now, consider $\H$'s equivalent matching $\M(\H)$ and learn an SMP predictor $\mu: \F \times \F^\prime \rightarrow \mathbb{R}$.
 Then, we have $(f, f^\prime) \in \M(\H)$ and $(\hat{f}, \hat{f}^\prime) \in \hat{\M}(\H)$. Hence, from eq. (3) in Definition 4, we have $\mu(f, f^\prime) \geq \mu(\hat{f}, \hat{f}^\prime)$. Now, if we define a derived BHP predictor $\varphi: \mathbb{B}(\F, \F^\prime) \rightarrow \mathbb{R}$ as $b \mapsto \mu(b\cap\V, b\cap\V^\prime)$, we get:
\begin{align*}
    \varphi(f \cup f^\prime) &= \mu((f\cup f^\prime)\cap\V, (f\cup f^\prime)\cap\V^\prime) = \mu(f, f^\prime)\nonumber\\
    &\geq \mu(\hat{f}, \hat{f}^\prime) = \mu((\hat{f}\cup \hat{f}^\prime)\cap\V, (\hat{f}\cup \hat{f}^\prime)\cap\V^\prime)\nonumber = \varphi(\hat{f} \cup \hat{f}^\prime)\nonumber\\
    \implies \varphi(f \cup f^\prime) &\geq \varphi(\hat{f} \cup \hat{f}^\prime).
\end{align*}
Hence, since $f$, $f^\prime$, $\hat{f}$, $\hat{f}^\prime$ were arbitrarily selected as per $\P$ and $\N$, the foregoing arguments make $\varphi$ satisfy the BHP condition given in eq. (2) from Definition 2.
\end{proof}

\section{Details on Experiments}
\label{app:exp}
\subsection{Data Description}
\label{app:exp:data_desc}
All datasets we have prepared are available on the following link:
\begin{center}
\texttt{\url{https://www.dropbox.com/s/cxykmi37695jlcw/catsetmat_data.zip?dl=0}}
\end{center}
Let us discuss the datasets one by one.
\subsubsection{TMDB Cast-Crew (\dataicc)}
We take a subset of movies from \textit{The Movie Database} (\textit{TMDB}; \texttt{\url{https://www.themoviedb.org/}}) available as a \textit{Kaggle} (\texttt{\url{https://www.kaggle.com/}}) dataset named \textit{TMDB 5000 Movie Dataset} (\texttt{\url{https://www.kaggle.com/tmdb/tmdb-movie-metadata}}).
The node sets considered here are those of \textit{movie-actors} and the \textit{movie-crew} (set of other important people involved in a movie, such as directors, producers, \textit{etc.}).
Every bipartite hyperedge contains a set of actors (the \textit{cast}; the left nodes) and a set of \textit{crew}-members (the right nodes) who participated in at least one movie.

\subsubsection{TMDb Cast-Keywords (\dataick)} TMDB also gives information about \textit{plot keywords} that hint at the content of a movie.
A bipartite hyperedge would contain a set of actors (the \textit{cast}; the left nodes) and a set of \textit{keywords} (the right nodes) related to least one movie.

\subsubsection{MAG ACM Authors-Keywords (\datamacm)}
This is a bibliographic network from \textit{Microsoft Academic Graph} (\textit{MAG})~\cite{sinha2015overview}, from which we choose those publications that appeared in conferences and journals associated with \textit{Association for Computing Machinery} (\textit{ACM}; \texttt{\url{https://www.acm.org/}}).
The two node sets used are those of \textit{authors} (the left node set) and (research-based) \textit{keywords} (the right node set), and each bipartite hyperedge contains a set of co-authors and a set of keywords at least one of their papers has been tagged with.

\noindent\textbf{Note}: The raw data for MAG is not available directly and it's a long process to collect portions of it. However, there used to be an online source that had official links to download the whole MAG dataset. We have an earlier version of the data downloaded with us, and this processing was made on that dataset. Nevertheless, the prepared data directly consumable by our code is available via the Dropbox link given earlier.

\subsection{Data Preparation}
\label{app:exp:data_prep}
\subsubsection{Processing Data}
We start with a raw dataset consisting of multiple entities, of which we choose two.
Then we fetch bipartite hyperedges, which are merely events occurring at a given point of time (e.g., a movie getting released, or a paper getting published, etc.); this gives us sets of higher-order co-occurrences of the two entities we chose.
Next, we apply a couple of filters to refine the data:
\begin{enumerate}
\item \textbf{Time filter}: We filter the data into events occurring at time $t \in [t_{min}, t_{max}]$.
\item \textbf{Occurrence frequency filter}: We only keep those nodes in the node-sets $\V$ and $\V^\prime$ that have their occurrence count within a particular range $[o_{min}, o_{max}]$.
\item \textbf{Size filter}: We only keep those bipartite hyperedges $f \cup f^\prime$ that have left and right hyperedge sizes (individually) in the range $|f|, |f^\prime| \in [s_{min}, s_{max}]$.
\end{enumerate}
Please note that the filters are applied sequentially, and the prepared data is stored as a two-column (left hyperedge, right hyperedge) CSV file.

\subsubsection{Negative Sampling}
\label{sec:exp:data_prep:neg_samp}
For the BHP problem, each bipartite hyperedge $f \cup f^\prime \in \B$ is considered to be a positive example ($\in \P$), and each non-hyperedge $f \cup f^\prime \in \hat\B := \mathbb{B}(\F, \F^\prime) \setminus \B$, a negative class sample.
But since the negative patterns are too many, we avoid the problem of class imbalance by \textit{negative class sampling}.
For this, we first pick two random hyperedges: one from left, ${f}\in \F$ and the other from right, ${f}^\prime \in \F^\prime$, and tag the pair as a negative class pattern ($\in \N$) if they aren't connected via a bipartite hyperedge (\textit{i.e.}, if ${f} \cup {f}^\prime \notin \B$).
\textit{We use a negative-to-positive ratio of $5{:}1$}.

\subsubsection{Train-Test Split}
We split the bipartite hyperedges and sampled bipartite non-hyperedges (sampled as mentioned in Section~\ref{sec:exp:data_prep:neg_samp} above) into train and test data using a $80{:}20$ train-test split ratio.
For \datamacm{}, we create validation data as well, using $60{:}20{:}20$ as train-validation-test proportions, and perform hyperparameter tuning on the validation data, as described in Section~\ref{sec:hyperparameter_tuning}.
For each dataset-algorithm combination, we perform \textit{five} experiments, performing a separate split and a separate negative sampling for each. This also introduces a standard deviation in our results, which we report in the main paper.

\subsection{Baselines}
\label{app:exp:baselines}
Barring the ``node2vec-based'' and the ``bipartite-graph-based'' baselines, we have explained all baselines in the main paper. Here is a detailed description of how both of them were created.
\subsubsection{Node2vec-based}
\label{app:exp:baselines:n2v}
For these set of baselines, we convert the bipartite hypergraph into a usual graph by expanding each bipartite hyperedge $b = f \cup f^\prime \in \B$ into a set of three types of edges:
    \begin{itemize}
        \item \textit{Cross-edges} $\E_{\times}(b) := f \times f^\prime$
        \item \textit{Left self-edges} $\E_\circ(b) := \{e \subseteq f: |e| = 2\}$
        \item \textit{Right self-edges} $\E_\circ^\prime(b) := \{e \subseteq f^\prime: |e| = 2\}$
    \end{itemize}
    Next, we embed each node $v$ into its node2vec~\cite{grover2016node2vec} features $X(v)$ and find cosine similarity between the incident nodes of each edge.
    Finally, we either take a mean or a min of that score and call it a prediction score.
    We use the same settings used in Hyper-SAGNN~\cite{zhang2020hyper} for their node2vec based baselines.
\begin{enumerate}
    \item \algoncme:
    For node2vec-cross-mean, we take $(v, v^\prime) \in \E_\times(b)$ and define the score to be:
    \begin{equation}
     n2v_{cross,mean}(b) := \frac{1}{|\E_\times(b)|}\sum_{(v, v^\prime) \in \E_\times(b)}{\!\!\!\!\!\!\!\!\!X(v)^TX(v^\prime)}.
    \end{equation}
    \item \algonfme: 
    \begin{equation}
     n2v_{full,mean}(b) := \frac{1}{|\E(b)|}\sum_{(u, v) \in \E(b)}{\!\!\!\!\!\!\!X(u)^TX(v)},
    \end{equation}
    where $\E(b) := \E_\cross(b) \cup \E_\circ(b) \cup \E_\circ^\prime(b)$ denotes the full set of edges.
    \item \algoncmi: 
    \begin{equation}
     n2v_{cross,min}(b) := \min_{(v, v^\prime) \in \E_\cross(b)}{\!\!\!\!\!\!X(v)^TX(v^\prime)}.
    \end{equation}
    \item \algonfmi: 
    \begin{equation}
     n2v_{full,min}(b) := \min_{(u, v) \in \E(b)}{\!\!\!\!\!X(u)^TX(v)}.
    \end{equation}
\end{enumerate}
\subsubsection{Bipartite-graph-based}
\label{app:exp:baselines:bip}
To create these baselines, we first convert our bipartite hypergraph $\H = (\V, \V^\prime, \F, \F^\prime, \B)$ into an induced bipartite graph $\G = (\V, \V^\prime, \E)$ by defining $\E$ as:
\[
\E := \{(v, v^\prime) \in \V \times \V^\prime \mid \exists b \in \B \text{ s.t. } \{v, v^\prime\} \subseteq b\},
\]
and then compute the bipartite versions of the Common Neighbor (CN) and the Adamic Adar (AA) scores as follows:
\[
CN(v, v^\prime) = \frac{LCN(v, v^\prime) + RCN(v, v^\prime)}{2},~ \text{and}~
AA(v, v^\prime) = \frac{LAA(v, v^\prime) + RAA(v, v^\prime)}{2},
\]
where ``L'' and ``R'' stand for left and right scores respectively. If $\Gamma$ denotes the neighbors of a node (or union of neighbors of a set of nodes), we can define the individual scores as (note that left nodes have neighbors in the right and vice-versa):
\[
LCN^\prime(v, v^\prime) = |(\Gamma(\Gamma(v)) \setminus\{v\}) \cap (\Gamma(v^\prime) \setminus\{v\})|
\]
\[
RCN(v, v^\prime) = |(\Gamma(\Gamma(v^\prime)) \setminus\{v^\prime\}) \cap (\Gamma(v) \setminus\{v^\prime\})|
\]
\[
LAA^\prime(v, v^\prime) = \sum_{w \in (\Gamma(\Gamma(v)) \setminus\{v\}) \cap (\Gamma(v^\prime) \setminus\{v\})}{\frac{1}{\log(1 + |\Gamma(w)|)}}
\]
\[
RAA(v, v^\prime) = \sum_{w^\prime \in (\Gamma(\Gamma(v^\prime)) \setminus\{v^\prime\}) \cap (\Gamma(v) \setminus\{v^\prime\})}{\frac{1}{\log(1 + |\Gamma(w^\prime|)}}
\]
We thus have $CN$ and $AA$ scores for each left-right vertex pair. Using this, we find the similarity scores for a left-right hyperedge pair $(f, f^\prime)$ as:
\[
\textsc{Algo}(f, f^\prime) = \textsc{Aggregate}(\{\textsc{Algo}(v, v^\prime) \mid (v, v^\prime) \in f \times f^\prime\}),\text{ and}
\]
where we use three choices for the \textsc{Aggregate} function and two for \textsc{Algo}, thereby forming the six baselines:
\begin{enumerate}
    \item \algolpminaamean: \textsc{Aggregate}=\textsc{Minimum}; \textsc{Algo}=AA
    \item \algolpmincnmean: \textsc{Aggregate}=\textsc{Minimum}; \textsc{Algo}=CN
    \item \algolpmaxaamean: \textsc{Aggregate}=\textsc{Maximum}; \textsc{Algo}=AA
    \item \algolpmaxcnmean: \textsc{Aggregate}=\textsc{Maximum}; \textsc{Algo}=CN
    \item \algolpavgaamean: \textsc{Aggregate}=\textsc{Average}; \textsc{Algo}=AA
    \item \algolpavgcnmean: \textsc{Aggregate}=\textsc{Average}; \textsc{Algo}=CN
\end{enumerate}

\subsection{Hyperparameter Settings}
\label{app:exp:hyp}
\subsubsection{Baselines}
There are no parameters for CN and AA scores for the bipartite graph based baselines.
For the node2vec-based baselines, we use embedding dimension as $64$, and other node2vec parameters are taken from Hyper-SAGNN.
For FSPool and HyperSAGNN, we take the latent dimension as $16$ and other parameters are as per Hyper-SAGNN. We tried this with other dimensions as well, but the best results were on this configuration.
\subsubsection{Hyperparameter tuning for \algocsm}
\label{sec:hyperparameter_tuning}
\begin{figure}[htb]
    \centering
    \includegraphics[width=0.6\linewidth]{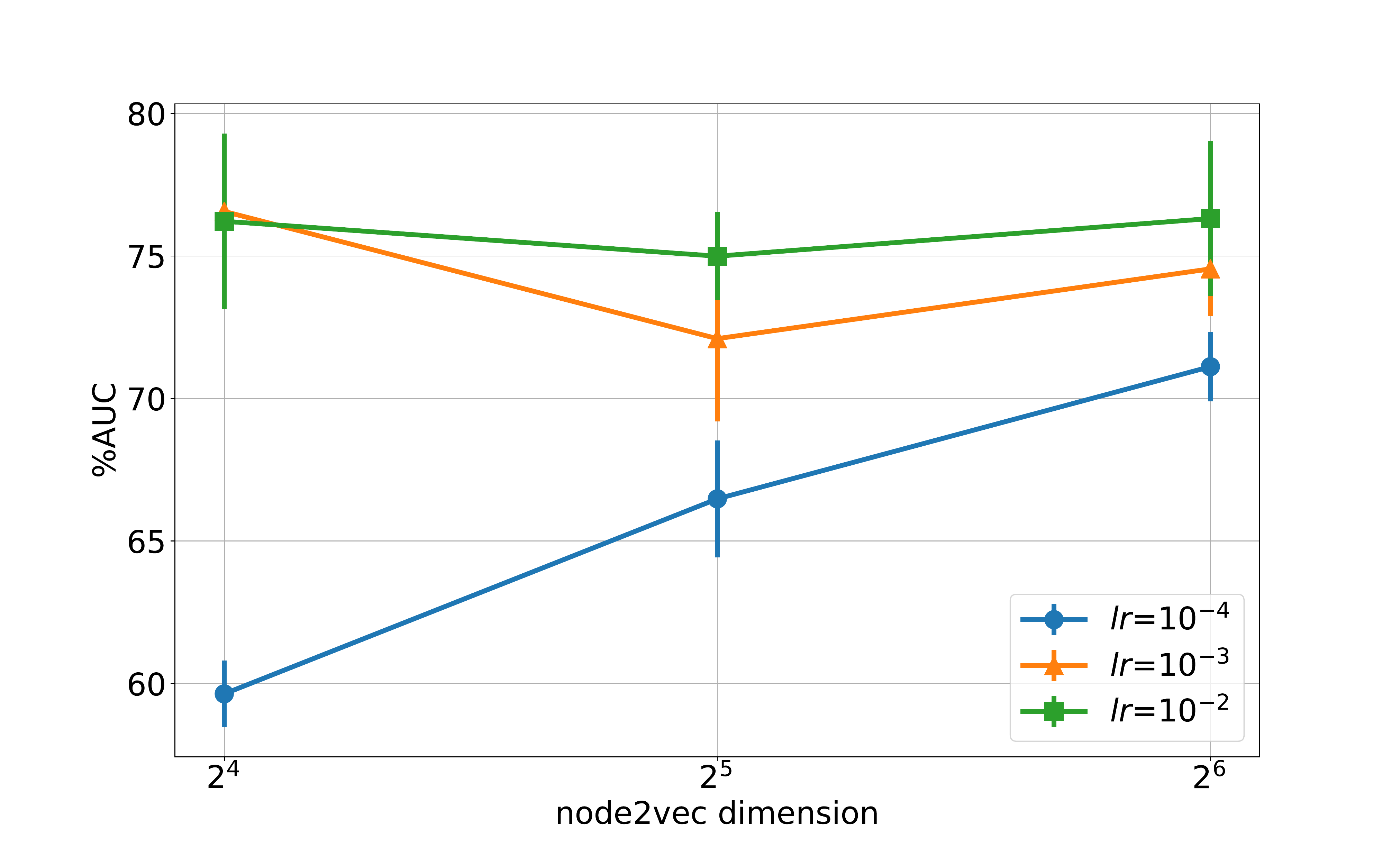}
    \caption{Tuning two hyperparameters on the dataset \datamacm{} for \algocsm: node2vec dimension $dim$ and learning rate $lr$.}
    \label{fig:hyperparameter_tuning}
\end{figure}

We fix some hyperparameters as per Hyper-SAGNN, and for the embedding dimension and the learning rate, we perform a hyperparameter tuning, as shown in Figure~\ref{fig:hyperparameter_tuning}, and then pick them to be $16$ and $0.001$ respectively.
We use the Adam optimization algorithm for our tasks.
The initial embeddings that we use in our model are from node2vec (an alternating random walk on nodes and hyperedges respectively), just as Hyper-SAGNN does.

